\documentclass[fleqn,10pt]{wlscirep}
\usepackage[utf8]{inputenc}
\usepackage[T1]{fontenc}
\usepackage{multirow}
\usepackage{adjustbox}

\title{A unified FLAIR hyperintensity segmentation model for various CNS tumor types and acquisition time points}

\author[1,*]{Mathilde Gajda Faanes}
\author[1]{David Bouget}
\author[2,3]{Asgeir S. Jakola}
\author[4]{Timothy R. Smith} 
\author[4]{Vasileios K. Kavouridis}
\author[5]{Francesco Latini}
\author[6]{Margret Jensdottir}
\author[7]{Peter Milos}
\author[8]{Henrietta Nittby Redebrandt}
\author[9]{Rickard L. Sjöberg}
\author[10]{Rupavathana Mahesparan}
\author[11]{Lars Kjelsberg Pedersen}
\author[12,13]{Ole Solheim}
\author[1,14]{Ingerid Reinertsen}

\affil[1]{Department of Health Research, SINTEF Digital, Trondheim, Norway}
\affil[2]{Department of Clinical Neuroscience, Institute of neuroscience and physiology, University of Gothenburg, Region Västragötaland, Gothenburg, Sweden}
\affil[3]{Department of Neurosurgery, Sahlgrenska University Hospital, Gothenburg, Sweden}
\affil[4]{Department of Neurosurgery, Brigham and Women’s Hospital, Harvard Medical School, Boston, MA, USA}
\affil[5]{Department Medical Sciences, Section of Neurosurgery, Uppsala University Hospital, Uppsala, Sweden}
\affil[6]{Department of Neurosurgery, Karolinska University Hospital, Stockholm, Sweden}
\affil[7]{Department of Neurosurgery, Linköping University Hospital, Linköping, Sweden}
\affil[8]{Department of Neurosurgery, Skåne University Hospital, Lund, Sweden}
\affil[9]{Department of Clinical Science, Umeå University, Umeå, Sweden}
\affil[10]{Department of Neurosurgery, Haukeland University Hospital, Bergen, Norway}
\affil[11]{Department of Neurosurgery, University Hospital of North Norway, Tromsø, Norway}
\affil[12]{Department of Neuromedicine and Movement Science, Norwegian University of Science and Technology, Trondheim, Norway}
\affil[13]{Department of Neurosurgery, St. Olavs University Hospital, Trondheim, Norway}
\affil[14]{Department of Circulation and Medical Imaging, Norwegian University of Science and Technology, Trondheim, Norway}

\affil[*]{mathilde.faanes@sintef.no}

\begin{abstract}
T2-weighted fluid-attenuated inversion recovery (FLAIR) magnetic resonance imaging (MRI) scans are important for diagnosis, treatment planning and monitoring of brain tumors. Depending on the brain tumor type, the FLAIR hyperintensity volume is an important measure to asses the tumor volume or surrounding edema, and an automatic segmentation of this would be useful in the clinic. In this study, around 5000 FLAIR images of various tumors types and acquisition time points from different centers were used to train a unified FLAIR hyperintensity segmentation model using an Attention U-Net architecture. The performance was compared against dataset specific models, and was validated on different tumor types, acquisition time points and against BraTS. The unified model achieved an average Dice score of 88.65\% for pre-operative meningiomas, 80.08\% for pre-operative metastasis, 90.92\% for pre-operative and 84.60\% for post-operative gliomas from BraTS, and 84.47\% for pre-operative and 61.27\% for post-operative lower grade gliomas. In addition, the results showed that the unified model achieved comparable segmentation performance to the dataset specific models on their respective datasets, and enables generalization across tumor types and acquisition time points, which facilitates the deployment in a clinical setting. The model is integrated into Raidionics, an open-source software for CNS tumor analysis. 

\end{abstract}
\begin{document}

\flushbottom
\maketitle
%
%
\thispagestyle{empty}

\section*{Introduction}
Central nervous system (CNS) tumors represent a diverse group of neoplasms and are classified by the World Health Organization (WHO) into over 160 different subtypes of brain and spinal cord tumors based on their histological and molecular characteristics \cite{louis_2021_2021}. The most common tumor entities are metastases, meningiomas, glioblastomas and lower grade gliomas (WHO 2-3). Magnetic resonance imaging (MRI) is a critical imaging modality for accurate diagnosis, treatment planning, evaluation of treatment responses, and longitudinal monitoring of brain tumors. Various MRI sequences are acquired to obtain information about the location, size, tumor type, and aggressiveness. T2-weighted (T2), T2-weighted fluid-attenuated inversion recovery (FLAIR) sequences and T1-weighted sequences before (T1w) and after application of a gadolinium-based contrast agent (T1c), is the core of the diagnostic gold standard \cite{weller_eano_2021}. Depending on the tumor type, the tumor may exhibit contrast enhancement, typically depicted on the T1c sequence. For contrast-enhancing tumors like metastases, glioblastoma, and meningioma, the tumor border and consequently the tumor volume is defined by the outer rim of T1 enhancement. For non- or less enhancing WHO grade 2-3 gliomas, the T2/FLAIR hyperintensity defines the tumor borders and tumor volumes. These tumors usually have no edema. However, for glioblastomas the FLAIR hyperintensity beyond the contrast-enhancing tumor has also gained clinical interest. This FLAIR-volume may represent a combination of peritumoral vasogenic edema and a gradient of infiltrating tumor cells, which can have importance for prognosis, defining the surgical target and evaluating treatment responses  \cite{wen_rano_2023}.For instance, the current RANO Resect framework states that patients with <5ml FLAIR hyperintensity volume and no residual contrast enhancement exhibit the best prognoses after surgery \cite{karschnia_prognostic_2023}. For meningiomas, the FLAIR-volume beyond the T1c enhanced tumor border is associated to WHO grade, i.e malignancy and for metastases the FLAIR hyperintensity volume represents vasogenic edema that may cause reversible neurological deficits. For any tumor undergoing radiotherapy, progressing FLAIR signals often represent treatment induced radiation gliosis. According to the new Response Assessment in Neuro-Oncology (RANO 2.0) criteria, both the contrast-enhancing and non-contrast-enhancing volumes in T1w and FLAIR MRI scans are important for treatment planning and prognostics for gliomas \cite{wen_rano_2023}. However, delineating these volumes manually is time consuming, with high inter- and intra-observer variability \cite{berntsen_volumetric_2020, visser_inter-rater_2019}. An accurate automatic segmentation of these hyperintesities hold potential to improve diagnostic accuracy, treatment planning, response evaluation, and prognostic evaluation, both pre- and post-operative for various tumor types. 

Through the datasets provided by the MICCAI BraTS challenges, among others, deep learning methods are already well studied for automatic delineations of brain tumors in MRI scans. With the rapid advancements in deep learning networks, different versions of the U-Net architecture \cite{ronneberger_u-net_2015}, such as nnU-Net \cite{isensee_nnu-net_2021}, Attention U-Net \cite{oktay_attention_2018}, and Swin U-Netr \cite{hatamizadeh_swin_2022}, have shown convincing results for medical image segmentation, like pre-operative glioma segmentation \cite{isensee_nnu-net_2021, hatamizadeh_swin_2022}. The BraTS challenge in 2023 provided a pre-operative glioma dataset with annotations of the enhancing tissue (ET), tumor core (TC), and whole tumor (WT) including the FLAIR hyperintensity volume. The winning team, using synthetic data augmentation, obtained a WT Dice score of 86.63 \% with T1w, T1c, T2 and FLAIR scans as input \cite{ferreira_how_2024}. In addition, the challenge provided a pre-operative meningioma dataset and pre-operative metastasis dataset with the same label types and input scans. The winning WT lesion-wise Dice scores were 85.6 \% and 60.2 \%, respectively. However, all four sequence types are needed as input, and the models are tumor type-specific for pre-operative scenarios only\cite{synapse_brats_2025-1}. In the BraTS challenge 2024, post-operative glioma segmentation was included for the first time with annotations of enhancing tissue (ET), non-enhancing tumor core (NETC), surrounding non-enhancing FLAIR hyperintensity (SNFH), and resection cavity (RC). The best lesion-wise SNFH Dice score obtained was 87.6 \% and the best lesion wise WT score obtained was 87.0 \% using the four MRI scans as input \cite{synapse_brats_2025}. However, using one unified model for FLAIR hyperintensity volume segmentation across multiple tumor types both before and after treatments would be more clinically applicable as the tissue diagnosis or the nature of the FLAIR hyperintensity volume is often not known.

In this work the aim was therefore to study automatic FLAIR hyperintensity segmentation models, with the following contributions: i) trained unified model, ii) thorough validation of segmentation of both FH and SNFH, iii) investigation of ability per tumor type and acquisition scan point, iv) benchmarked against BraTS, v) integrated and publicly available in the open software Raidionics \cite{bouget_preoperative_2022}.

\section*{Materials and Methods}
\label{sec:Data}
All methods in this study were carried out in accordance with regional and national regulations. Approvals were obtained from the ethical committee of Western Sweden (Dnr: 702-18), the Norwegian regional ethics committee (REK ref. 2013/1348 and 2019/510), and from the American Institutional Review Board (IRB 2023P001681).

\subsection*{Data}
\begin{figure}[!ht]
\centering
\includegraphics[width=0.95\textwidth]{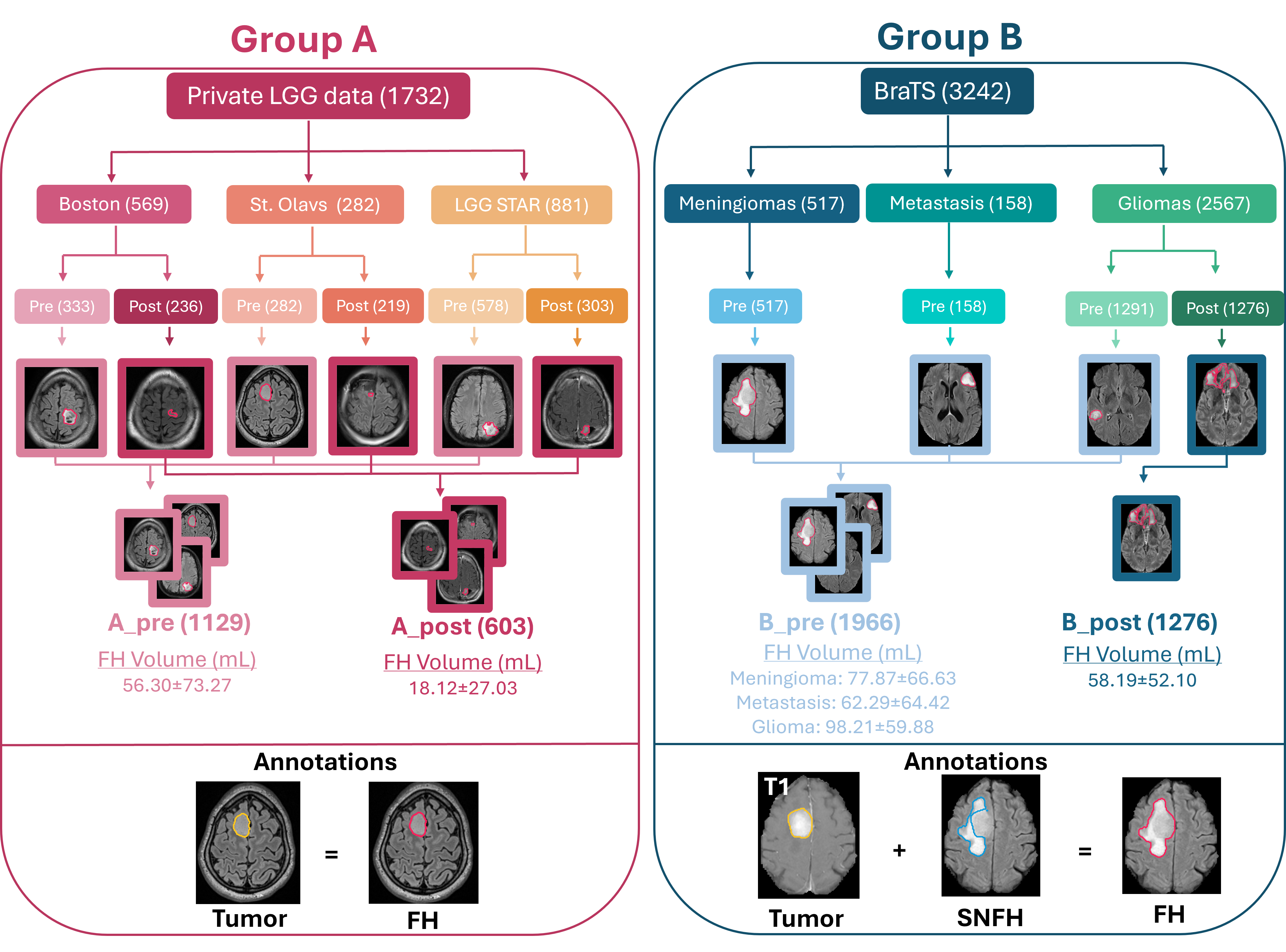}
\caption{Illustration of the data used in this project showing the data origin with the number of images in parenthesis, the annotation types, distribution into different subgroups for model training, and average FLAIR hyperintensity volumes with standard deviation. Group A represents the private LGG dataset of tumors classified as supratentorial diffuse WHO grade II glioma according to the 2007 or 2016 WHO classification system \cite{louis_2007_2007, louis_2016_2016}, with tumor annoations used as label for the FLAIR hyperintensity (FH) volume. Group B represents the BraTS data with annotations of the tumor and surrounding non-enhancing FLAIR hyperintensity (SNFH) volume, merged to produce the total FLAIR hyperintensity volume (FH). Group A has red-based colors, whereas group B has blue-based colors. Light colors represent pre-operative cases and dark colors represent post-operative cases.}
\label{fig:dataset}
\end{figure}
Data from multiple sources with various tumor types, volumes, and acquisition time points were used in this study. The data was divided into two groups, as presented in Figure \ref{fig:dataset}. Group A comprised a private dataset of patients with a supratentorial diffuse WHO grade II glioma classified according to the 2007 or 2016 WHO classification system \cite{louis_2007_2007, louis_2016_2016}. The data comes from Brigham and Women's Hospital (Boston, USA), St. Olavs hospital, Trondheim University Hospital (Trondheim, Norway), and from a Scandinavian multi-center cohort referred to as the LGG star group. It included annotated pre-operative and post-operative FLAIR scans, where 562 were taken up to 72 hours after surgery, 17 around one month after surgery, 19 around three moths after surgery, and 5 around 6 months after surgery. The average dimensions of the FLAIR scans were $[365\times 381 \times 91]$, ranging between $[156;1024] \times [192;1024] \times [6;512]$, with an average spacing of $[0.70 \times 0.70 \times 4.04]$, ranging between $[0.24;1.30] \times [0.24;1.30] \times [0.50;18.00]$ mm$^3$.

The datasets from group A contained annotations of tumor tissue which were used as label for the FLAIR hyperintensity volume, as seen in Figure \ref{fig:dataset}. No standardized annotation protocol was defined for the datasets from group A other than the overarching objective of performing tumor tissue segmentation. Other MRI sequences were also used during the annotation process to aid the interpretation and reduce disambiguation. For post-operative cases, the pre-operative images was also used to aid find residual tumor tissue. The annotations were obtained from multiple expert annotators from multiple centers. 

Group B comprised data publicly available from the BraTS challenges in 2023 and 2024 \cite{moawad_brain_2024, labella_asnr-miccai_2023, verdier_2024_2024, menze_multimodal_2015, baid_rsna-asnr-miccai_2021, bakas_advancing_2017}, including annotated FLAIR scans of pre-operative meningiomas, metastases, and gliomas, and post-operative gliomas acquired at different time points after surgery, where 578 were taken up to 72 hours after surgery, 484 around 1 month after surgery, 109 around 3 months after surgery, and 70 around 6 months after surgery. The grade of the gliomas were not specified, but they were mostly contrast-enhancing. No patient overlap exists between group A and B. The average dimensions of the FLAIR scans were $[217 \times 231 \times 166]$, ranging between $[182;240] \times [218;240] \times [155;182]$, with a spacing of $[1.0\times 1.0 \times 1.0]$ mm$^3$.

For group B, annotation of the tumor core (T) and surrounding non-enhancing FLAIR hyperintensity (SNFH), also called peritumoral edematous in BraTS 2023, were used. The tumor core comprises annotations of enhancing tissue and non-enhancing tumor core (called enhancing tumor and necrotic tissue in BraTS 2023). The tumor core annotations were obtained from hyperintensities in T1 and T1c images, while the SNFH annotations were obtained from the hyperintensities in the FLAIR scans around the tumor core, which could be edema, infiltrating tumor cells, and post-operative changes like ischemia \cite{moawad_brain_2024, labella_asnr-miccai_2023, verdier_2024_2024, menze_multimodal_2015, baid_rsna-asnr-miccai_2021, bakas_advancing_2017}. These two annotations were merged to form a label of the entire FLAIR hyperintensity volume, as illustrated in Figure \ref{fig:dataset}. 

To reduce annotation noise in the datasets, cases with a FH volume between 0 and 0.1\,mL were excluded, corresponding to annotations of up to 100\,voxels. Cases with a volume of 0 mL were kept as true negative samples. In addition, a similar threshold 0.1\,mL was used over model predictions to determine positiveness (i.e., above threshold) or negativeness (i.e., below threshold) of the sample. All cases were positive, except of 81 cases from the post-operative images from group A.

\subsection*{Methods}
In this section, the experimental setup, preprocessing, model training, and evaluation will be described. The pipeline for the data preparation and training of the segmentation models are illustrated in Figure \ref{fig:pipeline}. 

\subsubsection*{Experimental setup and naming convention}

\begin{figure}[!ht]
\centering
\includegraphics[width=0.99\textwidth]{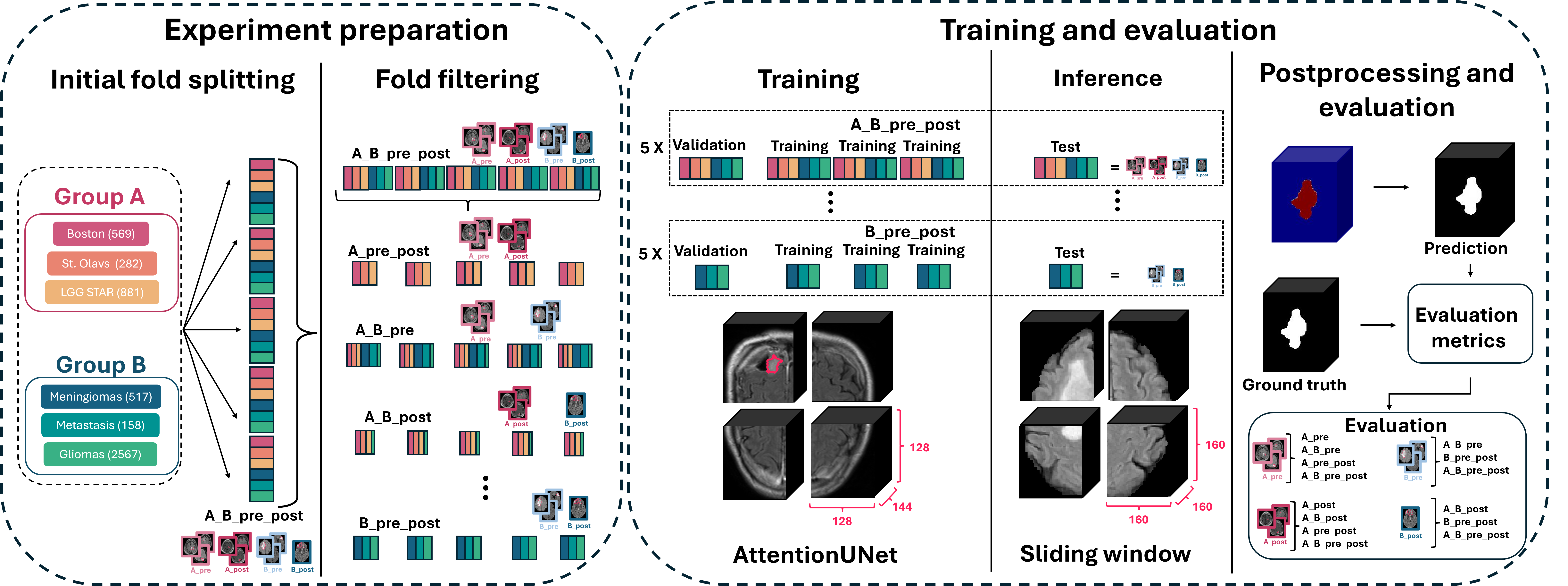}
\caption{Illustration of the experiment preparation and the pipeline for training and evaluating segmentation models. An overview of the fold splitting, where the data were evenly split with respect to tumor type, source and size (represented by colors), and of how different subgroups of the data were used as input for training and evaluation, is presented. Note that each model was evaluated on each subgroup it was trained on, and that since the fold splits were kept fixed, the test folds are the same for each subgroup for all models.}
\label{fig:pipeline}
\end{figure}

In this study, different configurations of the datasets were used for training and evaluation of the models, as illustrated in Figure \ref{fig:pipeline}. All data was initially split into five evenly-distributed folds to perform three-way five-fold cross-validation. For each training, one fold was used as test set, one as validation set, and the rest as training set. To ensure a fair comparison, the data was split patient-wise with an even distribution of the data source and tumor type in each fold, and ensuring both small and larger tumors from each source in each fold. Multiple models were trained using different subgroups of the data, and the models were evaluated on test sets from their respective subgroups. The same training strategy was used for all models, enforcing consistent fold splits across all experiments, ensuring comparability across the different configurations and ensuring an equal test set for all models evaluated on the same subgroup.


Depending on the segmentation target, data with different label-types were used. To train FH segmentation models, the original tumor annotations from group A were directly used, while both SNFH and tumor annotation masks from group B were merged (cf. Figure \ref{fig:dataset}). To train SNFH models, i.e., the surrounding FLAIR hyperintensity volume excluding contrast-enhancing tumor tissue, only the original SNFH annotation masks from group B were used.

Each model configuration was named descriptively by prepending the dataset group (e.g., A, B, or A\_B for both) and appending the acquisition time point (e.g., pre, post, or pre\_post for both). For further validation studies based off the tumor type, another prefix was prepended where Gli, Met, and Men respectively represent glioma, metastasis, and meningioma.

\subsubsection*{Preprocessing}
For preprocessing, all data were resampled to an isotropic voxel spacing of 1\,mm$^3$ with a first-order spline interpolation. To reduce background, tight cropping around the patient's head was performed. In addition, intensity clipping was performed within the range [0,99.5]\%, and zero-mean normalization of all nonzero values was performed.

\subsubsection*{Architecture design and training strategy}
In this study, all models were trained with the Attention U-Net architecture \cite{oktay_attention_2018}. The architecture was used with 5 levels with filter sizes of $[16,32,64,256,512]$, instance normalization, a dropout rate of $0.2$, and an input size of $128\times128\times144$\,voxels. A combination of Dice and Cross-Entropy was used as loss function with AdamW as optimizer combined with an annealing scheduler and an initial learning rate of $5e{-4}$. The sigmoid function was used for the final activation layer, and the models were trained over $800$ epochs with a batch size of 16 and a 2 step gradient accumulation strategy, giving an effective batch size of 32. 

The models were trained with data augmentation. First, a random crop of $128\times128\times144$ voxels was applied to each input to match the network input size. Next, both geometric transformations and intensity-based transformations were applied. For geometric transformations, rotation within the range $[-20^{\circ}, 20^{\circ}]$, flipping along each axis, zoom scaling up to 15 \%, and translation of up to 20 \% of the axis dimension was applied, each with a 50 \% probability. Intensity scaling and shifting up to 10 \%, gaussian noise addition, gamma contrast adjustments in the range [0.5,2.0], and patch inversion or dropout with patch sizes of $10\times10\times10$ voxels and up to 75 elements, were applied with a 50 \% probability for the intensity-based transformations.


\subsubsection*{Inference and postprocessing}
For inference, the inference patch size was set to $160\times160\times160$ voxels, and a sliding-window approach with 50 \% overlap between the patches was performed. 

As postprocessing, a binary brain mask was applied to the predictions, ensuring only the prediction probabilities inside the brain were kept. In addition, prediction probabilities with a volume lower than $0.05\,mL^3$ or not visible in two consecutive 2D slices, were removed to reduce noise.

\subsection*{Evaluation metrics}
\label{subsec:measandmet}
To compare and quantify model performance, different metrics were computed between the ground truth volume and a binary representation of the predicted probability maps obtained after inference with the trained models. The binary representation was computed for ten different probability thresholds between $[0,1]$. The threshold providing the best scores were used to report the evaluation metric scores for all test samples for each fold. 

To asses various aspects of performance, metrics focusing on classification capability, segmentation accuracy, and capturing clinically relevant outcomes were chosen.

\paragraph{Classification:} Since the majority of cases were positive, except 81 cases in A\_post, the evaluation focused solely on the models' ability to correctly identify a finding when present. Detection rate (also referred to as image-wise recall) was used, as the percentage of true positive cases (TP) among the total number of positive cases (P). A voxel-wise Dice score of $0.1\%$ between model prediction and ground truth was required for a positive case to be considered true positive.

\paragraph{Segmentation:} To evaluate segmentation performance, when the model correctly detected something, object-wise metrics were computed on the true positive samples. They describe a model's ability to detect all components in case of multiple volumes of interest. A pairing strategy combined with a connected components approach was used to associate the ground truth and the model predictions' components. From this, Dice score, recall, precision and 95th percentile Hausdorff distance (HD95), were computed and reported as mean $\pm$ standard deviation.

\paragraph{Clinical use:} Regarding clinically-oriented metrics, reporting volume difference has potential for diagnosis and comparison with clinical guidelines. To evaluate this, the number of oversegmented and undersegmented cases and their respective volume differences were calculated among the true positive samples and reported as median [Interquartile Range].

\section*{Experiments}
The aim of this study was to develop and validate a unified FLAIR hyperintensity volume segmentation model that can be used for various brain tumor types, acquisition time points, and also for SNFH segmentation. In the following, the experiments performed are described.  

\subsection*{Evaluation of training set configurations}
First, the unified model (A\_B\_pre\_post) was trained using all subgroups of the data presented in Figure \ref{fig:dataset}, thus A\_pre, A\_post, B\_pre, and B\_post, with FLAIR hyperintensity (FH) as label. To evaluate the performance of the unified model, dataset-specific models were trained using different configurations of the subgroups and compared to the unified model. The models were evaluated on each subgroup they were trained on, and since the fold splits were kept fixed, the test folds were the same for each subgroup for all models, ensuring a fair comparison, as shown in Figure \ref{fig:pipeline}. Only the segmentation metrics were used to compare the models. 

The different configurations were selected based on a balance between training efficiency and the specific insights we aimed to obtain. Six dataset-specific models were trained with FLAIR hyperintensity (FH) as label using different combinations of A\_pre, A\_post, B\_pre and B\_post to obtain these models following the naming convention presented: A\_pre, A\_post, A\_pre\_post, A\_B\_pre, A\_B\_post, and B\_pre\_post.

In addition, for certain tumor types, segmenting only the surrounding non-enhancing FLAIR hyperintensity (SNFH) volume, without the tumor core, might be of interest. Another dataset-specific model was thus trained with SNFH as label using only B\_pre and B\_post since group A do not have SNFH labels, called B\_pre\_post$^*$. To obtain this volume for the unified model, the tumor volumes were subtracted from the unified model's predictions and evaluation metrics were computed against the original SNFH labels from group B. To ensure a fair comparison, the tumor volumes were also subtracted from the predictions of this dataset-specific model. $^*$ represent the tumor volume subtraction and that the model is for SNFH.

\subsection*{Generalizability study}
Next, the unified model, trained on all data, was more thoroughly validated on different tumor types, targets, volumes, and for different pre- and post-operative scenarios using the classification, segmentation, and clinical metrics presented.


\subsection*{Benchmarking against BraTS}
Lastly, the unified model was compared against the best scores from the BraTS Meningioma challenge 2023, BraTS Brain Metastasis challenge 2023, BraTS Adult Glioma challenge 2023, and BraTS Adult Glioma Post Treatment challenge 2024, thus ensuring proper benchmarking against current state-of-the-art models. Segmentation of the Whole Tumor (WT) from BraTS corresponds to the FLAIR hyperintensity volume (FH), and segmentation of the surrounding non-enhancing FLAIR hyperintensity volume (SNFH) is the same. The test sets used in the challenges are not publicly available, therefore the scores were collected from the official leaderboards available on Synapse showing the results on the validation sets \cite{synapse_brats_2025-1,synapse_brats_2025}. 

\section*{Results}
The training, inference, and validation processes were implemented in Pyhton 3.11 using \texttt{PyTorch} v2.4.0, \texttt{PyTorch Lightning} v2.4.0 , and \texttt{MONAI} v1.4.0 \cite{cardoso_monai_2022}. An Intel Core Processor (Broadwell, no TSX, IBRS) CPU with $16$ cores, $64$GB of RAM, Tesla A40 ($46$ GB) or A100 ($80$GB) dedicated GPUs, and NVMe hard-drives was used for the experiments.


\subsection*{Evaluation of training set configurations}
\begin{table}[!h]
\caption{FLAIR hyperintensity (FH) volume and surrounding FLAIR hyperintensity (SNFH) volume segmentation performance summary for models trained with different configurations of the dataset. Each model is evaluated on a test set from each subgroup it has been train on. The unified model (A\_B\_pre\_post), has been train and tested on all subgroups and is highlighted in bold. $^*$ represent the tumor volume subtraction for SNFH segmentation.}
\centering
\adjustbox{max width=\textwidth}{
\begin{tabular}{l|cc|cccc}
 & & & \multicolumn{4}{c}{Object-wise} \tabularnewline
Model & Test set & Target & Dice & Recall & Precision&HD95 \tabularnewline
\hline
A\_pre  & A\_pre & FH& $84.52\pm13.59$ & $85.82\pm15.32$ & $86.28\pm14.36$ & $04.95\pm05.42$ \tabularnewline
A\_pre\_post & A\_pre  &FH& $84.06\pm13.55$  & $85.26\pm15.87$ & $86.12\pm14.27$ & $05.28\pm06.09$ \tabularnewline
A\_B\_pre & A\_pre  &FH& $84.90\pm13.15$ & $87.66\pm14.14$ & $85.17\pm14.87$ & $04.88\pm05.71$ \tabularnewline
\textbf{A\_B\_pre\_post }& \textbf{A\_pre}  & \textbf{FH}&$\mathbf{84.47\pm13.32}$ & $\mathbf{88.79\pm14.70}$ & $\mathbf{83.48\pm14.64}$ & $\mathbf{04.97\pm05.52}$ \tabularnewline
\hline
A\_B\_pre & B\_pre & FH&$89.25\pm11.48$ & $87.80\pm14.29$ & $93.03\pm08.52$ & $03.91\pm05.08$ \tabularnewline
B\_pre\_post & B\_pre & FH&$89.26\pm11.44$ & $89.22\pm13.68$ & $91.47\pm09.96$ & $03.96\pm05.03$ \tabularnewline
\textbf{A\_B\_pre\_post} & \textbf{B\_pre} & \textbf{FH}&$\mathbf{89.51\pm10.60}$ & $\mathbf{89.77\pm12.99}$ & $\mathbf{91.32\pm09.26}$ & $\mathbf{03.88\pm04.70}$ \tabularnewline
\hline
A\_post &  A\_post &  FH&$61.37\pm23.54$ & $74.42\pm22.73$ & $61.15\pm27.16$& $10.42\pm08.42$ \tabularnewline
A\_pre\_post& A\_post  &FH& $62.19\pm24.07$ & $75.31\pm23.71$ & $62.19\pm27.27$ & $09.99\pm08.25$ \tabularnewline
A\_B\_post &  A\_post  & FH&$61.26\pm23.96$ & $73.02\pm24.79$ & $61.69\pm27.00$ & $10.15\pm07.98$ \tabularnewline
\textbf{A\_B\_pre\_post}&  \textbf{A\_post} & \textbf{FH} &$\mathbf{61.27\pm24.35}$ & $\mathbf{78.26\pm24.55}$ & $\mathbf{58.60\pm26.63}$ & $\mathbf{10.26\pm08.44}$ \tabularnewline
\hline
A\_B\_post & B\_post & FH& $84.41\pm12.02$ & $85.27\pm12.88$ & $86.57\pm13.06$ & $04.77\pm06.23$ \tabularnewline
B\_pre\_post & B\_post & FH&$84.75\pm12.14$ & $86.35\pm12.65$ & $86.21\pm13.35$ & $04.76\pm06.10$ \tabularnewline
\textbf{A\_B\_pre\_post} & \textbf{B\_post} &\textbf{FH}& $\mathbf{84.60\pm11.78}$ & $\mathbf{86.55\pm12.42}$ & $\mathbf{85.62\pm13.19}$ &$\mathbf{04.86\pm06.47}$ \tabularnewline
\hline
B\_pre\_post$^*$  & B\_pre  & SNFH& $81.02\pm17.75$ & $81.96\pm18.59$ & $85.08\pm17.26$ & $05.43\pm06.66$ \tabularnewline
\textbf{A\_B\_pre\_post$^*$} & \textbf{B\_pre}   & \textbf{SNFH}& $\mathbf{83.40\pm16.87}$ & $\mathbf{86.96\pm16.06}$ & $\mathbf{84.98\pm17.47}$ & $\mathbf{05.20\pm07.05}$ \tabularnewline
\hline
B\_pre\_post$^*$  & B\_post  &SNFH&  $83.23\pm13.19$ & $85.05\pm14.55$ & $85.36\pm13.21$ & $05.06\pm06.83$ \tabularnewline
\textbf{A\_B\_pre\_post$^*$}  & \textbf{B\_post} &\textbf{SNFH}& $\mathbf{83.94\pm12.76}$ & $\mathbf{85.43\pm13.44}$ & $\mathbf{85.96\pm13.76}$ &$\mathbf{04.95\pm06.87}$ \tabularnewline
\end{tabular}
}
\label{tab:all_models}
\end{table}
Segmentation performance for the various dataset configurations are summarized in Table \ref{tab:all_models}. For each test set, and for both targets, the dataset-specific models and the unified model achieved similar Dice and HD95 scores. Similarly, recall and precision scores are comparable for all models on all test sets. The A\_post dataset is an exception, where the unified model (A\_B\_pre\_post) obtained higher recall and lower precision than the dataset-specific models. In addition, there is a variation in performance across the different test set, varying from an average Dice score of $61.27\%$ for the A\_post dataset, to $89.51\%$ for the B\_pre dataset for the unified model.

\subsection*{Generalizability study}


\begin{table}[!h]
\caption{Segmentation performance with the FLAIR hyperintensity (FH) and the surrounding non-enhancing FLAIR hyperintensity (SNFH) volumes as target for the unified (A\_B\_pre\_post) model, on meningiomas (Men), metastasis (Met), and gliomas (Gli) from dataset A and B, and for both pre- and post-operative acquisition time points. Detection rate is calculated on the positive cases, and the other metrics are calculated on only true positive cases.}
\centering
\adjustbox{max width=\textwidth}{
\begin{tabular}{lc|c|cccc|cc|cc}
&  &  &\multicolumn{4}{c|}{} &\multicolumn{2}{c|}{Oversegmentation} &\multicolumn{2}{c}{Undersegmentation}   \tabularnewline
Test set  & Target & Detection rate & Dice & Recall & Precision & HD95 & $\Delta$ (mL)   &\# Samples & $\Delta$ (mL) & \# Samples \tabularnewline
\hline
Men\_B\_pre & FH  & 98.07 &  $88.65\pm12.47$ & $89.00\pm15.38$ & $90.52\pm09.26$ & $04.05\pm04.75$ &3.30 [1.34-6.89] &272& 3.24 [1.39-8.04] &235 \tabularnewline
Met\_B\_pre & FH  & 92.41 &$80.08\pm13.75$ & $80.81\pm16.90$ & $84.73\pm13.61$ & $04.33\pm03.94$ & 4.11 [1.01-10.25] &71& 4.87 [1.45-10.64] &75\tabularnewline
Gli\_B\_pre & FH  &99.85 &  $90.92\pm08.56$ & $91.09\pm10.79$ & $92.39\pm08.21$ & $03.76\pm04.84$& 3.74 [1.33-7.35] &569& 5.16 [2.02-10.34] &720 \tabularnewline
Gli\_B\_post & FH & 99.84 & $84.60\pm11.78$ & $86.55\pm12.42$ & $85.62\pm13.19$ & $04.86\pm06.47$& 3.21 [1.20-6.96] &624& 3.01 [1.34-6.18] &650 \tabularnewline
Gli\_A\_pre & FH & 98.05 & $84.47\pm13.32$ & $88.79\pm14.70$ & $83.48\pm14.64$ & $04.97\pm05.52$ &3.14 [1.23-8.28] &605& 3.26 [1.10-7.80] &502 \tabularnewline
Gli\_A\_post & FH & 90.61 & $61.27\pm24.35$ & $78.26\pm24.55$ & $58.60\pm26.63$ & $10.26\pm08.44$ & 3.97 [1.43-9.60] &331& 2.39 [1.00-4.56] &142 \tabularnewline
\hline
Men\_B\_pre & SNFH & 96.91 &  $78.96\pm23.13$ & $90.83\pm16.73$ & $76.20\pm24.33$ & $08.05\pm09.61$ & 3.42 [1.48-7.51] &409& 1.12 [0.39-2.95] &92 \tabularnewline
Met\_B\_pre & SNFH & 92.41 & $75.57\pm19.28$ & $77.77\pm20.87$ & $80.22\pm19.31$ & $04.24\pm03.78$& 3.18 [1.07-9.34] &79& 3.13 [1.06-9.42] &67 \tabularnewline
Gli\_B\_pre & SNFH & 99.85& $86.02\pm12.37$ & $86.50\pm14.54$ & $88.94\pm11.62$ & $04.24\pm05.75$ & 3.39 [1.31-7.00] &617& 4.09 [1.61-8.94] &672\tabularnewline
Gli\_B\_post & SNFH & 99.84 & $83.94\pm12.76$ & $85.43\pm13.44$ & $85.96\pm13.76$ &$04.95\pm06.87$& 3.24 [1.24-7.21] &664& 2.29 [1.05-4.84] &610 \tabularnewline
\end{tabular}
}
\label{tab: tumor types}
\end{table}

\begin{figure}[!ht]
\centering
\includegraphics[width=0.9\textwidth]{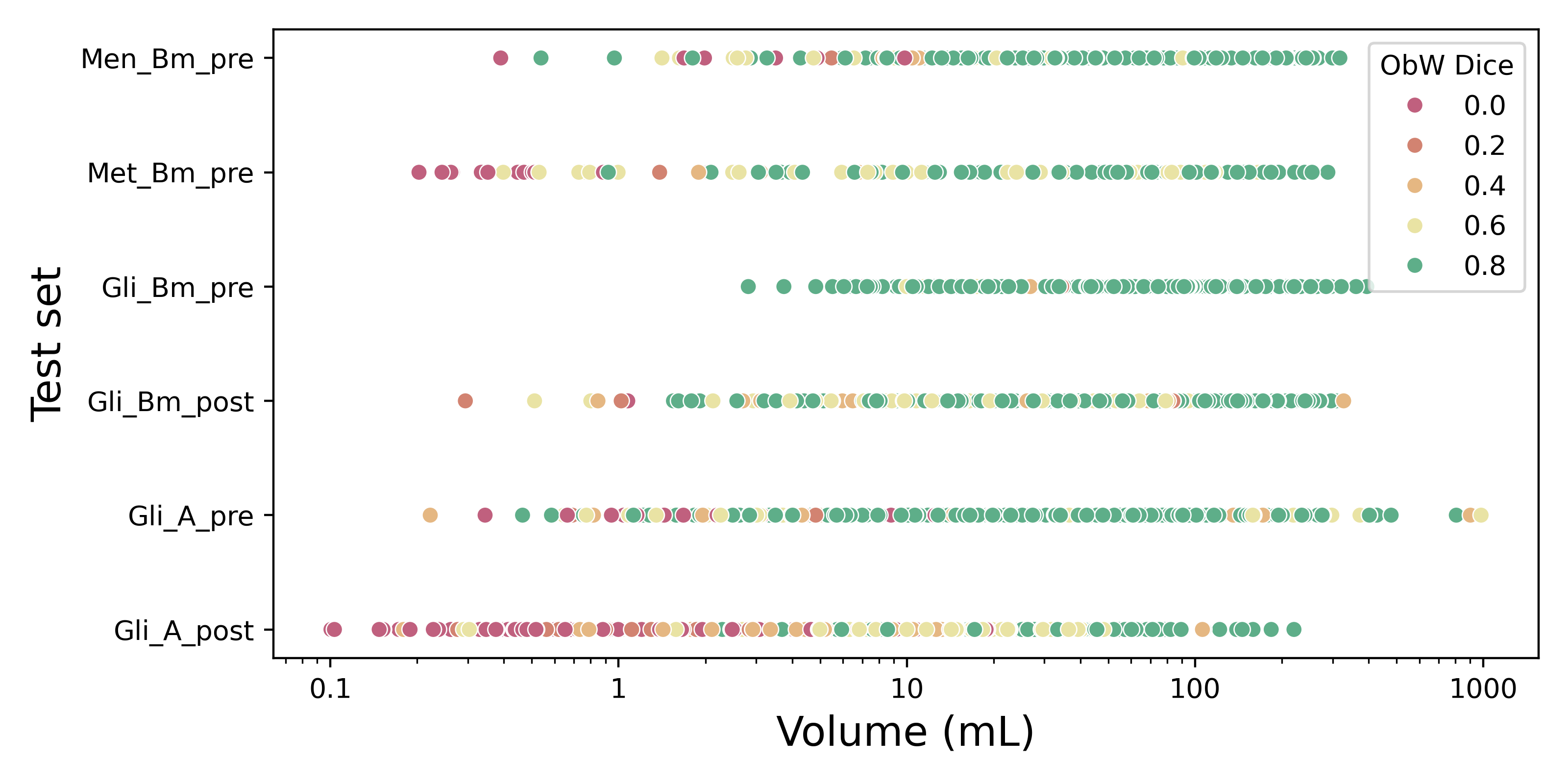}
\caption{Scatterplot showing the object-wice Dice scores for the unified model (A\_B\_pre\_post) on different test sets grouped by brain tumor type pre- and post-operatively along the Y-axis and different volumes with a logarithmic scale along the X-axis. All test cases are shown, including false negative samples.}
\label{fig:type_volume_dice}
\end{figure}

\begin{figure}[!ht]
\centering
\includegraphics[width=0.9\textwidth]{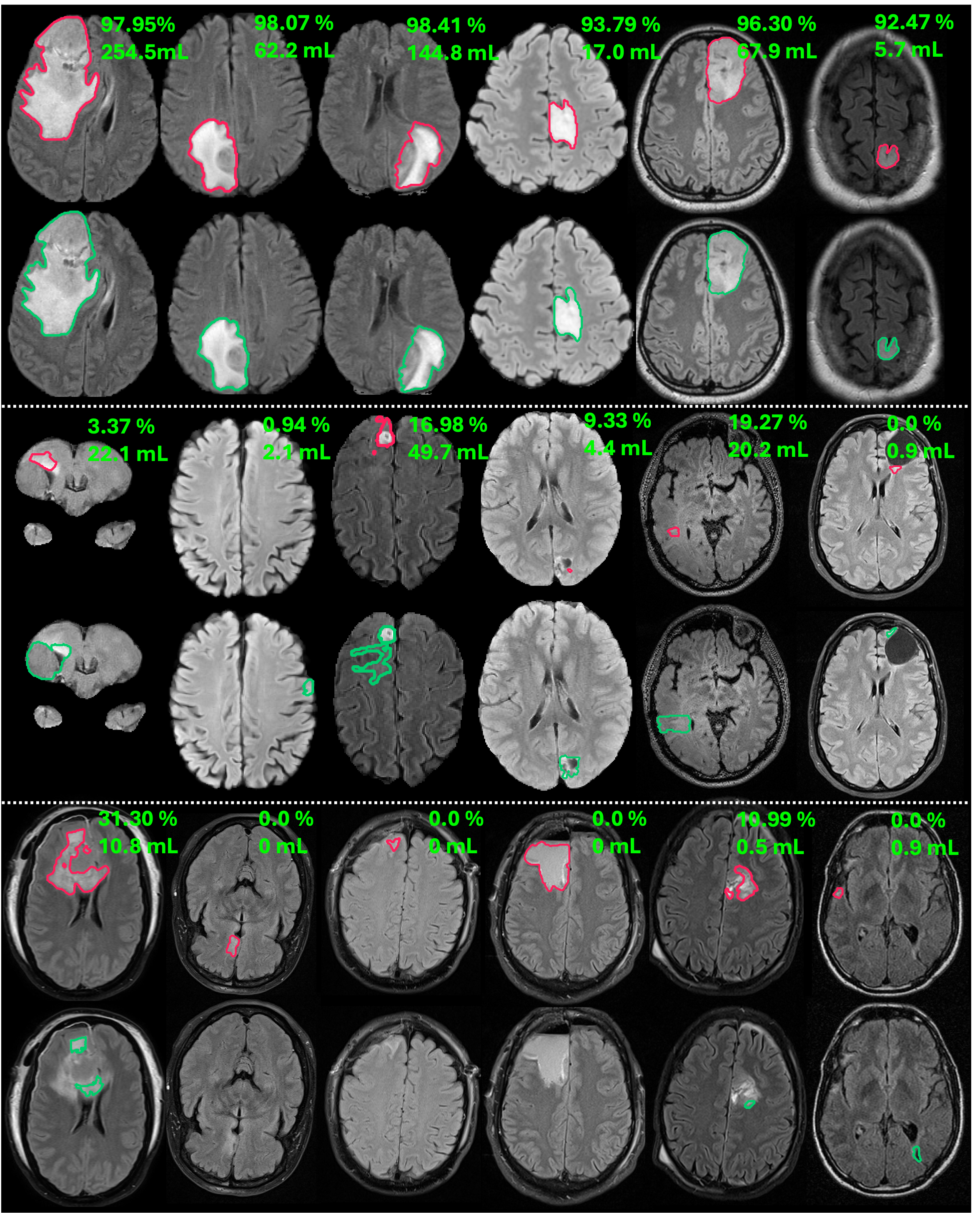}
\caption{Examples of predictions in pink and the ground truth in green. The top row and the second row shows cases with high and low Dice scores, respectively. From left to rights samples from: Men\_B\_pre, Met\_B\_pre, Gli\_B\_pre, Gli\_B\_post, Gli\_A\_pre and Gli\_A\_post. The last row shows examples of Gli\_A\_post with low scores. }
\label{fig:pred_ex}
\end{figure}

The performance of the unified model (A\_B\_pre\_post) on different tumor types, acquisition time points, and with FH or SNFH as target are summarized in Table \ref{tab: tumor types}. It shows a difference in performance across the groups. For both targets, the unified model scored best on Gli\_B\_pre, obtaining the best average object-wise Dice scores with high detection rates, high and similar recall and precision scores, and low HD95 scores. The amount of over- and undersegmented cases are equally distributed, with small volume differences. Moreover, the unified model achieved good scores on Men\_B\_pre and Gli\_B\_post and slightly worse on Met\_B\_pre with a lower detection rate, for both FH and SNFH. For FH segmentation of dataset A, the unified model achieved good scores for Gli\_A\_pre, but struggled with Gli\_A\_post were it obtained the lowest and most variable scores, with a high recall, but low precision, and obtained a high number of oversegmented cases compared to the undersegmented cases. In addition, the unified model generally scored lower for SNFH than for FH. Moreover, Figure \ref{fig:type_volume_dice} shows the relationship between Dice score and volume distributions for the different test sets comprising of different tumor types and acquisition time points with FH as target. It shows that the cases with smaller volumes generally give lower Dice scores.

Table \ref{tab:times} shows the performance on FH segmentation on scans acquired before (Pre), early after (Early post), and later after (Late post) surgery, which includes the scans acquired around 1 to 6 months after surgery (Post1, Post3, and Post 6). Among the 81 negative cases of the post-operative gliomas from group A, 55 were considered as false positive from Early post, 1 from Post 1 and 1 from Post 6. Some examples of the predictions, including examples of false positive and false negative cases, can be seen in Figure \ref{fig:pred_ex}, which shows one row of examples with high and one with low Dice scores for all tumor types and acquisition time points, and one row of extra examples of the Gli\_A\_post cases. For the lower row, some ground truth masks are debatable (number 1 and 2), or missing some lesions (number 6), and in others the surgical cavity with hyperintense FLAIR-signal as seen early postoperatively, is included in the automatic segmentation masks (number 3, 4, and 5).

\begin{table}[!h]
\caption{Segmentation performance for the unified model on dataset A, B and the entire dataset grouped by different acquisition time points, where "Pre" were acquired pre-operatively, "Early post" were acquired right after surgery, and "Late post" were acquired around 1-6 months after surgery. For the entire dataset "Late post" is also split in "Post 1","Post 3" and "Post 6", which indicate the rounded-up number of months post-surgery at which the MRI scan is taken. Note that meningiomas and metastasis do not have post-operative images, and are thus only in the pre-operative group.}
\adjustbox{max width=\textwidth}{
\begin{tabular}{l|c|cccc|cc|cc}
&   &\multicolumn{4}{c|}{} &\multicolumn{2}{c|}{Oversegmentation} &\multicolumn{2}{c}{Undersegmentation}   \tabularnewline
Acquisition time point& Detection rate & Dice & Recall & Precision & HD95 & $\Delta$ (mL)   &\# Samples & $\Delta$ (mL) & \# Samples \tabularnewline
\hline
\textbf{Dataset A} &&&&&&&&& \tabularnewline
Pre&98.05 &$\mathbf{84.47\pm13.32}$ & $\mathbf{88.79\pm14.70}$ & $\mathbf{83.48\pm14.64}$ & $\mathbf{04.97\pm05.52}$ & 3.14 [1.23-8.28] &605& 3.26 [1.10-7.80] &502 \tabularnewline
Early post &90.72 & $60.60\pm24.54$ & $77.96\pm24.75$ & $57.99\pm26.85$ & $10.39\pm08.46$ & 3.76 [1.32-9.42] &310& 2.39 [1.02-4.56] &130 \tabularnewline
Late post &89.19 & $70.38\pm21.13$&$80.33\pm23.04$&$70.94\pm22.88$& $08.53\pm08.09$ & 5.87 [4.28-12.44] &21& 2.52 [0.78-4.37] &12 \tabularnewline
\hline
\hline
\textbf{Dataset B} &&&&&&&&& \tabularnewline
Pre&98.78 & $\mathbf{89.51\pm10.60}$ & $\mathbf{89.77\pm12.99}$ & $\mathbf{91.32\pm09.26}$ & $\mathbf{03.88\pm04.70}$ & 3.58 [1.33-7.32] &912& 4.64 [1.75-10.03] &1030 \tabularnewline
Early post&99.84 & $83.59\pm12.92$ & $86.17\pm12.97$ & $83.90\pm14.76$ &$05.01\pm06.45$ & 3.36 [1.35-7.28] &316& 2.99 [1.39-5.57] &296 \tabularnewline
Late post &99.85 & $85.60\pm10.70$&$85.9\pm12.18$&$88.3\pm11.2$& $04.73\pm06.49$ &3.01 [1.07-6.80] &308& 3.13 [1.20-6.44] &354 \tabularnewline
\hline
\hline
\textbf{Dataset A and B} &&&&&&&&& \tabularnewline

Pre  & 98.51 &  $87.68\pm11.92$ & $89.42\pm13.65$ & $88.47\pm12.12$ & $04.27\pm05.08$ & 3.41 [1.30-7.63] &1517& 4.21 [1.50-9.46] &1532 \tabularnewline
Early post &95.81 & $73.98\pm21.86$ & $82.72\pm19.28$ & $73.07\pm24.34$& $07.20\pm07.81$ &3.57 [1.33-7.78] &626& 2.77 [1.24-5.45] &426 \tabularnewline
Late post  & 99.29 & $84.88\pm11.83$&$85.64\pm12.93$&$87.48\pm12.54$ & $04.91\pm06.54$ & 3.21 [1.09-7.41] &329& 3.06 [1.20-6.37] &366 \tabularnewline
\hline
Post 1  & 99.80 & $84.20\pm11.53$ & $86.08\pm12.33$ & $85.78\pm12.65$&$04.91\pm06.34$ &3.22 [1.07-6.81] &242& 3.17 [1.19-6.32] &257\tabularnewline
Post 3  & 97.62 & $85.76\pm11.37$ & $88.13\pm12.58$ & $86.47\pm12.88$ & $05.38\pm07.70$ & 3.39 [1.24-8.02] &55& 2.73 [1.41-6.73] &68
\tabularnewline
Post 6  &  98.64 & $87.27\pm11.43$ & $88.84\pm11.54$ & $88.46\pm13.23$ & $04.12\pm05.76$ &2.09 [1.24-7.07] &32& 2.23 [1.03-6.22] &41 \tabularnewline

\end{tabular}
}
\label{tab:times}
\end{table}



\subsection*{Benchmarking against BraTS}

The performance of our unified model on our test sets compared to the best scores on the validation sets from the BraTS challenges are presented in Table \ref{tab:brats}. It shows that our model obtain comparable results for FH segmentation of pre-operative gliomas, meningiomas and metastases, and for FH and SNFH segmentation of post-operative gliomas on our test set compared to the validation set scores from the BraTS challenge. 


\begin{table}[h]
\centering
\caption{Our unified model (A\_B\_pre\_post) compared to the results on the validation set for the BraTS challenges 2023 and 2024. Whole tumor (WT) corresponds to the FLAIR hyperintensity (FH) volume. $^*$ represent the tumor subtraction from the predictions to get the SNFH volume.}
\adjustbox{max width=\textwidth}{
\begin{tabular}{l||ccccc||c}
 & & & & & & \multicolumn{1}{c}{Object/lesion-wise} \tabularnewline
Model & Tumor type & Target  & \# Samples & \# Inputs & Test/Val& Dice \tabularnewline
\hline

Our & Meningioma & FH/WT & 517 & 1 & Test & $88.65\pm12.47$ \tabularnewline
BraTS 2023 & Meningioma & FH/WT & 141 & 4& Val & $85.6$ \tabularnewline
\hline
Our & Metastasis & FH/WT &158 & 1& Test & $80.08\pm13.75$ \tabularnewline
BraTS 2023  & Metastasis & FH/WT & 31& 4& Val & $60.2$ \tabularnewline
\hline
Our & Glioma pre & FH/WT & 1291 & 1&Test & $90.92\pm08.56$ \tabularnewline
BraTS 2023 & Glioma pre & FH/WT & 219 & 4&Val & $91$ \tabularnewline
\hline
Our & Glioma post & FH/WT & 1276 & 1& Test & $84.57\pm11.92$ \tabularnewline
BraTS 2024 & Glioma post & FH/WT & 188 & 4&Val & $87.6$  \tabularnewline
\hline
Our$^*$ & Glioma post & SNFH & 1276 &  1& Test & $84.60\pm11.78$  \tabularnewline
BraTS 2024  & Glioma post & SNFH & 188 & 4&  Val& $87.2$  \tabularnewline
\end{tabular}
}
\label{tab:brats}
\end{table}

\section*{Discussion}

This study aimed to develop and validate a robust unified FLAIR hyperintensity segmentation model for various brain tumors and acquisition time points. First, dataset-specific models were trained and compared against the unified model. Then the unified model was validated on different tumor types and acquisition time points, and the model was used with tumor subtraction to evaluate the performance on the surrounding non-enhancing FLAIR hyperintensity volume. Finally, the model was benchmarked against the top performing models trained on the BraTS dataset.

The dataset used in this study comprises a large and diverse patient population, covering all major CNS tumors, including meningiomas, metastases, and various WHO grades of gliomas. However, the key imitation in this study is that the data used originates from different sources without a common annotation protocol for the FLAIR hyperintensity volume. For the BraTS annotations in group B, the original labels were of the SNFH and the tumor volume which were merged to get the whole FLAIR hyperintensity volume. This could result in differences in the ground truth mask compared with those that would be obtained if the FLAIR hyperintensity volume were annotated directly. For Group A, comprising low grade gliomas, the original labels were of the tumor volume that were used for the whole FLAIR hyperintensity volume. For low grade gliomas, the entire pre-operative FLAIR hyperintensity volume is associated with the whole tumor volume because these tumors have no or little edema. Nevertheless, some edema can occur during treatment, and the post-operative tumor volume could thus be smaller than the post-operative FLAIR hyperintensity volume. In addition, the lack of a standardized annotation protocol in such a multi-centric dataset surely introduced noise and unwanted variations in the annotations. For example, in case of multiple tumors, all tumors or just the one planned for resection could be annotated. Also, there could be variations if the entire tumor volume was annotated or just the volume that was planned resected. This inconsistent and possible smaller ground truths could thus induce noise in the training, and is not ideal for evaluation, where the performance may be better than what the scores indicate. Re-annotating all data with the same annotation protocol is time consuming and requires expert-knowledge and was therefore not feasible. In addition, there is a high inter-rater variability for FLAIR hyperintensity segmentation \cite{bo_intra-rater_2017}, which may diminish the noise created by the uncertainties in the ground truths.  

The unified model (A\_B\_pre\_post) achieved comparable segmentation performance to the dataset-specific models, with both FH and SNFH as target, as shown in Table \ref{tab:all_models}. Even though some of the dataset-specific models offer minor improvements in performance on their respective dataset, the unified model aligns good segmentation performance with versatility enabling generalization across tumor types and acquisition time points. This simplifies deployment in a clinical setting because there is no need to choose a specific model based on tumor type or scan time, which is especially beneficial when the tumor type is still unknown. A unified model is thus more desirable for this application and was selected for a more thorough analysis and validation.

The validation of the unified model on different tumor types, pre- and post-operative showed generally good scores with both FH and SNFH as targets, however there were some differences across the groups. Figure \ref{fig:type_volume_dice} indicates that the Dice score is largely affected by the volume. Across all groups, smaller volumes generally corresponded to lower Dice scores, while larger volumes tended to yield higher scores. It seems that for all groups, cases with a volume lower than around 1 mL are challenging to segment. In addition, the lower performing groups had overall smaller volumes than the higher performing groups. From Figure \ref{fig:dataset} it can be seen that there is a large difference in the average FLAIR hyperintensity volume of the different tumor types and acquisition time points, where the post-operative gliomas from group A have the lowest average volume of $18.12\pm27.03$ mL, and the pre-operative gliomas from group B have the largest average volume of $98.12\pm59.88$ mL. This could explain why the pre-operative gliomas from group B obtained the highest Dice scores and why the post-operative gliomas from group A obtained the worse score. In addition, it can be seen that the unified models fails to segment most of the cases under 1 mL for the meningiomas and post-operative gliomas from group B, which could explain the lower detection rate for these groups. Moreover, it can explain why the SNFH scores are generally lower than the FH scores, because the volumes are generally lower. This is a well-known phenomenon in machine learning since smaller objects tend to disappear before the final, most abstract, feature maps. In addition, small objects give a high class imbalance which is a challenge in the learning process. Also, Dice score is highly sensitive to errors in small structures, and Dice loss is suboptimal as it penalizes more heavily very small segmentation differences  \cite{reinke_common_2023}. This indicates thus that the volume affects the Dice scores more than the tumor type and acquisition time points. In addition, group A has a higher average slice thickness which affects the quality of the images and might affect the performance of the model and the calculation of the evaluation metrics. This could also explain why the results are generally better for the data from group B.

However, FH appears to be particularly difficult to segment in dataset Gli\_A\_post, as performance is generally lower that over the other datasets, with similar structure volume. Several factors can cause such performance drop, starting by the tendency of the unified model to perform oversegmentation over this subgroup, as seen in Table \ref{tab: tumor types}. According to Figure \ref{fig:pred_ex}, the model tend, in some cases, to segment something when the ground truth is empty, to segment a larger volume than the ground truth, or to segment another area than the ground truth. However, some clearly visible FH regions were not included in the ground truth, which can be ascribed to the lack of standardized annotation protocol as previously discussed. The performance on this specific subgroup could thus be better than reflected in the reported metrics, and the unified model's exposure to datasets with labels approximating better the true FH volume could be an advantage for post-operative non-contrast-enhancing glioma segmentation, enabling the model to segment FH regions more accurately. In addition, the inter-rater variability of manual FH segmentation can be substantial \cite{bo_intra-rater_2017}, with some cases with a debatable ground truth. For segmentation of diffuse FLAIR hyperintensity volumes, high precision (i.e. reproducibility) may be at least as important as high accuracy to increase reproducibility of image assessments in clinical practice, clinical trials, and in treatment registries.

In addition, post-operative scans are particularly difficult to segment as the edges are often unclear and disturbed by swelling, ischemia, blood products, treatment induced proteins, hemostatic agents, and other cavity-fluids that could increase FLAIR signals. This is supported by the results from Table \ref{tab:times}, indicating higher scores over pre-operative scans than post-operative, for both groups. Post-operatively, early scans taken up to 72 hours after surgery obtained the lowest scores, while performance improves for late post-operative scans, acquired 1 to 6 months after surgery. The reason for this could be that the disturbing factors that occur right after treatment have had time to settle and are less pronounced in later scans, giving clearer edges and an easier segmentation task. This can also be seen in Figure \ref{fig:pred_ex}, which shows that it is more difficult to identify blood products, cavity fluid or FLAIR hyperintensity caused by edema or tumor tissue in early post-operative scans. The model also wrongly segments the cavity fluid in some of the examples, which can be explain by that proteins in surgical cavities increase FLAIR-signals.


Compared to BraTS, Table \ref{tab:brats} shows that our models obtained similar scores on all tasks, and higher score over the metastasis category. However, since the scores from BraTS are from the validation set, the BraTS scores might be higher due to the model being validated on this set, in contrast to our results that were derived from a test set never seen by the model. The validation sets are also smaller than our test sets, which makes them more sensitive to randomness. Also, the BraTS models are given more information with four MRI sequences as input, T1w, T1c, T2 and FLAIR, whereas our model uses only FLAIR scans. In addition, the implementation of the pairing strategy to compute the object-wise scores in our work, and lesion-wise scores in BraTS may slightly differ, which can impact the results. This could explain why our unified model obtained an object-wise Dice score of $80.08$\% on the metastasis subgroup where multifocal tumors are common, compared to $60.2$\% for BraTS. This could also be explained by our choice to report performance over true positives only, whereby the detection rate was lower than for the other groups. Our results are thus not directly comparable to BraTS, nevertheless such benchmarking still provides a useful indication of our model's performance, and shows that our models obtain good scores on the different BraTS tasks.


Regarding clinical usage, the unified model has achieved promising results and could be useful in a clinical context to assess the FLAIR hyperintensity volume, both in pre-operative and post-operative scenarios. The median volume differences are low compared to the average FH volumes, supporting the use of the segmented volumes for assessing clinical guidelines. For non-contrast-enhancing tumors, such as low grade gliomas, this volume is important to assess the tumor volume \cite{wen_rano_2023}. However, a limitation in this study is that some clinicians prefer to use T2 scans instead of FLAIR scans to assess the tumor volume. For this study, more FLAIR images were available than T2 scans, and only FLAIR scans were thus used for this study. For future work,  it would be interesting to validate our models against annotations in T2 scans directly. For contrast-enhancing tumors (such as meningioma, glioblastoma and metastasis), it could be interesting to only analyze the surrounding FLAIR hyperintensity volume. Since the contrast-enhancing tumor volumes often are known, either from manual annotations or from high-performing segmentation software, such as Raidionics \cite{bouget_preoperative_2022}, the tumor volume can be subtracted from the predictions from the unified model to obtain this volume. As seen, the unified model performs well, and it also achieved better results than the dataset-specific model trained using only the SNFH labels presented in Figure \ref{fig:dataset}. The surrounding FLAIR hyperintensity volume is especially important after surgery for the contrast-enhancing tumors to assess edema which is linked to survival \cite{karschnia_prognostic_2023, wen_rano_2023}. However, more thorough clinical validation is needed before clinical implementation \cite{majewska_prognostic_2025, kommers_glioblastoma_2021}. An important part of the clinical validation would be to define what the FLAIR hyperintensity volume is. There are biological differences depending on the tumor type, and there are also differences depending on the center, neurosurgeon, and if the goal is to get the removable FLAIR hyperintensity volume, or the entire volume, especially for the non- or less enhancing WHO grade 2-3 gliomas, which must be considered before a more thorough clinical validation.

Even though the unified models achieved good performance, there is room for further improvements, especially for cases with a volume smaller than around 1 mL. First, as discussed, new labels can be acquired with a standardized annotation protocol to reduce the noise in the annotations. This can be obtained more rapidly by manually adjusting predictions from the unified model, or by interactive active learning. Also, the dataset should be expanded to obtain more cases without any FLAIR hyperintensities to balance the dataset. Images from true negatives could be used since there is often some remaining FLAIR hyperintensity volumes in the post-operative images, especially for contrast-enhancing tumors which often still have edema after surgery. Other training strategies more optimal for small structures can also be tested, such as multi-scale approaches or different loss functions. In addition, the model can be improved further by providing multiple MRI sequences as inputs and extending to multi-class segmentation of the FLAIR hyperintensity volume, the contrast-enhancing volume and cavity. If some MRI sequences are missing, synthetic MRI sequences can be generated using denoising diffusion models. In addition, denoising diffusion models could be used to translate 2.5D MR scans, with high slice thickness, into 3D scans with appropriate resolution, avoiding the added blurriness from naive resampling techniques. 

\section*{Conclusion}


This research showed that our unified FLAIR hyperintensity segmentation model aligns good segmentation performance across different brain tumor types and acquisition time points with good generalization. The unified model simplifies deployment in a clinical setting and has potential for clinical use. The unified model is integrated and publicly available in the open software Raidionics \cite{bouget_preoperative_2022}.



\section*{Acknowledgements}
Data were processed in digital labs at HUNT Cloud, Norwegian University of Science and Technology, Trondheim, Norway. 


\section*{Funding}
M.G.F, D.B, I.R, and O.S are partly funded by the Norwegian National Research Center for Minimally Invasive and Image-Guided Diagnostics and Therapy. The LGG STAR group, as part of the study, was financed by grants from the Swedish state under the agreement between the Swedish government and the county councils, the ALF-agreement (ASJ; ALFGBG-1006089)

\section*{Author contributions statement}
A.S.J, T.R.S, V.K.K, F.L, M.J, P.M, H.N.R, R.L.S, R.M., L.K.P, and O.S collected and curated the data; M.G.F conceived the experiments; M.G.F conducted the experiments; M.G.F analyzed the results; M.G.F wrote the original draft; D.B., M.G.F, O.S, and I.R reviewed and edited the manuscript; I.R and O.S acquired the funding. All authors reviewed the manuscript.

\section*{Additional information}
\textbf{Accession codes} the Raidionics environment with all related information is available at \url{https://github.com/raidionics}. More specifically, all trained models can be accessed at \url{https://github.com/raidionics/Raidionics-models/releases/tag/1.3.1}, the Raidionics software can be found at \url{https://github.com/raidionics/Raidionics}. Finally, the source code used to compute the validation metrics is available at \url{ https://github.com/dbouget/validation_metrics_computation}

\textbf{Competing interests} The authors declare no competing interests. The funders had no role in the design of the study; in the collection, analyses, or interpretation of data; in the writing of the manuscript; or in the decision to publish the results.

\bibliography{references}

\end{document}